\documentclass[conference]{IEEEtran}

\usepackage{cite}
\usepackage{graphicx}
\usepackage{booktabs}
\usepackage{amsmath}
\usepackage{array}
\usepackage{tabularx}
\usepackage{url}
\usepackage{tikz}
\usetikzlibrary{positioning,arrows.meta}
\usepackage{standalone}
\usepackage[hidelinks]{hyperref}

\newcolumntype{Y}{>{\raggedright\arraybackslash}X}

\hypersetup{
  pdftitle={Nedimo: An Evidence-Grounded Multilingual Pipeline for AI-Generated Sleep Stories},
  pdfauthor={Georgios Makridis and Stefanos Kampouris},
  pdfsubject={Evidence-grounded multilingual generation and synthesis of sleep stories},
  pdfkeywords={generative AI, sleep stories, text-to-speech, safety guardrails, sleep onset}
}

\title{Nedimo: An Evidence-Grounded Multilingual\\
Pipeline for AI-Generated Sleep Stories}

\author{%
\IEEEauthorblockN{Georgios Makridis and Stefanos Kampouris}
\IEEEauthorblockA{Department of Digital Systems\\
University of Piraeus\\
Piraeus, Greece\\
\textit{Web application and project website:} \href{https://nedimo.com}{nedimo.com}\\
\textit{Mobile apps:} \href{https://play.google.com/store/apps/details?id=com.nedimo}{Google Play}, \href{https://apps.apple.com/app/id6803620912}{App Store}}}

\begin{document}

\maketitle

\begin{abstract}
Generative audio systems can produce large libraries of personalized bedtime content, but ordinary media objectives---retention, completion, and novelty---are poorly aligned with the goal of supporting a transition toward sleep. Direct evidence for narrated bedtime audio in adults is mixed, and a recent active-controlled trial did not support such audio as a standalone treatment for sleep disturbance. We translate this uncertain evidence into conservative engineering requirements and present a multilingual pipeline for generating and narrating low-arousal sleep stories. The implemented prototype separates deterministic planning from large language model generation, represents content and delivery parameters explicitly, shapes each script through a four-phase attention taper, validates structural and lexical constraints, performs a bounded error-aware retry, and quarantines unresolved outputs. Provider-agnostic text-to-speech synthesis is followed by paragraph-aware joining, two-pass loudness normalization, and a publication gate that re-measures the saved, encoded file against fixed loudness, loudness-range, and true-peak limits. The product data model records stop position, replay, early abandonment, and completion while treating these events as ambiguous behavioral proxies rather than clinical outcomes. We contribute an evidence-to-requirement mapping, a traceable guardrail architecture with explicit acceptance thresholds, technical evaluation criteria, and a prospective protocol for testing sleep-onset effects. No clinical efficacy is claimed; the system is positioned as a relaxation and attentional-support tool whose benefit and possible reliance must be evaluated empirically.
\end{abstract}


\section{Introduction}
Audio stories are increasingly offered as pre-sleep content, yet their design objective differs from that of conventional spoken media. For an entertainment service, finishing a story and requesting another episode are desirable outcomes. For a sleep-onset aid, continuous engagement to the end may instead indicate that the listener remained awake. Optimizing ordinary completion metrics can therefore work against the intended use.

The empirical basis also requires restraint. An early audiobook comparator did not improve sleep quality over control~\cite{harmat2008}; small app and intensive-care studies reported partial or context-specific signals~\cite{kirk2022,economides2023,mermer2024}; and a 2026 randomized trial of 495 working adults found that bedtime stories did not outperform an active digital control on its primary sleep-disturbance outcome~\cite{vazzaz2026}. At the same time, in a self-selected online survey of 1,136 Swedish adults, respondents with sleep-onset insomnia symptoms more often reported using bedtime audiobooks to reduce worry, and daily listeners more often reported returning to the same book~\cite{berglund2026}. These results motivate a system that is usable and testable without claiming that narration is a treatment for insomnia.

This paper describes the design and implementation of such a system in the \href{https://nedimo.com}{Nedimo web application and project website}~\cite{nedimo} and its Android and iOS apps~\cite{nedimo-play,nedimo-appstore}. The work addresses three questions: how sleep-related evidence can be translated into controllable generation parameters; how those constraints can be enforced across text generation, synthesis, and delivery; and how the system can be evaluated without equating media engagement with sleep.

The principal contributions are:
\begin{itemize}
\item an evidence-to-requirement mapping for adult sleep-story generation;
\item a deterministic, multilingual orchestration pipeline with a parameterized narrative contract and four-phase attention taper;
\item layered structural, lexical, and acoustic guardrails with stated acceptance thresholds, bounded retry, and quarantine; and
\item a three-level evaluation framework separating technical conformance, product telemetry, and clinical sleep outcomes.
\end{itemize}

\section{Evidence Base and Design Requirements}

\subsection{Narration, attention, and adult sleep}
The most plausible role of a story is attentional redirection: it can occupy enough cognitive capacity to interrupt rumination while avoiding the uncertainty and emotional activation of ordinary narrative. Harvey and Payne found that people with insomnia who elaborated an interesting, engaging, pleasant, and relaxing image reported shorter sleep-onset latency and less uncomfortable pre-sleep thought than participants assigned general distraction or no instruction~\cite{harvey2002}. This finding argues against a completely flat story. It instead suggests an attentional middle: concrete and gently absorbing, but without consequential goals.

Direct bedtime-audio evidence remains heterogeneous (Table~\ref{tab:direct-evidence}). Harmat et al. compared relaxing music, an audiobook, and control in 94 students; music improved subjective sleep quality, whereas the audiobook did not~\cite{harmat2008}. A 40-person crossover study found actigraphy-estimated sleep-efficiency and pre-sleep heart-rate-variability changes across sleepcasts, music, and mindfulness, but only mindfulness improved subjective PSQI scores~\cite{kirk2022}. A 300-person pilot supported the feasibility of app-delivered stories and music but lacked the stronger active-control evidence needed for efficacy claims~\cite{economides2023}. Mermer and Arslan reported favorable perceived sleep and physiological measures after a 30-minute audiobook-plus-music session in 56 intensive-care patients, a setting that limits generalization~\cite{mermer2024}. Most decisively, the larger USleep trial found only a very small, nonsignificant primary effect for bedtime stories relative to digital control ($g=0.12$, 95\% CI $[-0.13,0.37]$)~\cite{vazzaz2026}.

\subsection{Sound and delivery}
Adjacent music evidence is more developed than story evidence. A Cochrane review found improvement in subjective sleep quality, while effects on objective measures were less certain~\cite{jespersen2022}. Sleep-associated music is generally softer and slower than music overall, but exhibits substantial individual and subgroup variation~\cite{scarratt2023}. Reviews of white and pink noise are heterogeneous, so an ambient layer should remain optional rather than be presented as universally effective~\cite{capezuti2022}. WHO environmental guidance recommends very low continuous bedroom noise; this is a boundary for avoiding disturbance, not a validated audiobook playback dose~\cite{who2009}. Likewise, no fixed carrier, tuning, binaural, or ``brainwave'' frequency has been established as an effective property of narrated sleep audio.

Delivery may matter independently of story content. In an adjacent experiment, 35 participants heard a soothing narrated voice, a robotic voice, or silence after a standardized stress-induction task; salivary cortisol fell faster in the soothing-voice group, whereas electrodermal recovery did not differ between groups~\cite{saskovets2025}. This is a stress-recovery finding, not a sleep outcome, and we treat it only as a rationale for testing calm, stable prosody. Screen-off playback is preferable because evening exposure to light-emitting readers can delay sleep and circadian timing~\cite{chang2015}. These findings motivate stable prosody, low dynamic variation, a timer, a gradual fade, and disabled autoplay; they do not establish a universal narrator pitch or speech rate.

\subsection{Constrained generation and speech synthesis}
Large language models have been used to generate personalized, age-appropriate fairy tales for children, together with prompts for accompanying illustrations~\cite{makridis2024fairyland}. Sleep narration for adults adds constraints that such storytelling systems do not target: low arousal, a controlled attentional taper, and a duration budget. Prompting alone does not guarantee schema-valid output. Structured-output benchmarks show that even capable language models can violate nontrivial schemas, supporting the use of an external parser and validator~\cite{lu2025}. Precise length following is also unreliable in zero-shot generation; target adjustment, filtering, and automated revision improve compliance~\cite{retkowski2025}. For synthesis, SSML provides a provider-independent representation for pronunciation, pauses, rate, pitch, and volume, although provider behavior remains implementation specific~\cite{ssml2010}. Modern neural TTS systems also expose one-to-many variation in duration, pitch, and rhythm~\cite{kim2021}; consequently, textual correctness alone cannot guarantee acoustic conformance. The final waveform must be measured. We use loudness, loudness-range, and true-peak descriptors consistent with EBU R~128 terminology, without assuming that its broadcast target is itself an appropriate bedtime level~\cite{ebu128}.

\subsection{Evidence-to-requirement translation}
Tables~\ref{tab:direct-evidence} and~\ref{tab:requirements} summarize how the literature is operationalized. These are conservative design hypotheses, not proven therapeutic ingredients.

\begin{table*}[!t]
\caption{Direct and Closely Related Adult Bedtime-Audio Evidence}
\label{tab:direct-evidence}
\centering
\scriptsize
\begin{tabularx}{\textwidth}{p{2.3cm}p{3.2cm}p{3.4cm}Y}
\toprule
Study & Design and sample & Intervention & Implication for the present system \\
\midrule
Harmat et al.~\cite{harmat2008} & Controlled study, $n=94$ students & 45 min/night for three weeks: music, audiobook, or control & Spoken content is not intrinsically sleep promoting; content and delivery require explicit design. \\
Kirk et al.~\cite{kirk2022} & Randomized within-person crossover, $n=40$ & Sleepcasts, music, mindfulness, and no-intervention weeks & Supports feasibility and possible physiological effects, but subjective benefit was not specific to stories. \\
Economides et al.~\cite{economides2023} & Multi-arm pilot, $n=300$ working adults & App-based stories or music over four weeks & Supports scalable delivery and retention; pilot findings require active-controlled confirmation. \\
Mermer and Arslan~\cite{mermer2024} & Quasi-experimental intensive-care study, $n=56$ & One 30-min audiobook-plus-music session & Provides context-specific positive evidence but does not isolate narration or generalize to home use. \\
Vazzaz et al.~\cite{vazzaz2026} & Four-arm active-controlled RCT, $n=495$ & Stories, sounds, skills, or digital control for four weeks & No superiority on the primary outcome; motivates conservative claims and controlled evaluation. \\
Berglund et al.~\cite{berglund2026} & Cross-sectional self-selected online survey, $n=1{,}136$ Swedish adults & Self-reported bedtime audiobook use & Describes respondents' motives, familiarity, and perceived reliance; cannot establish prevalence or efficacy. \\
\bottomrule
\end{tabularx}
\end{table*}

\begin{table*}[!t]
\caption{Evidence-to-Requirement Mapping}
\label{tab:requirements}
\centering
\scriptsize
\begin{tabularx}{\textwidth}{p{4.2cm}p{5.0cm}Y}
\toprule
Evidence or risk & Design requirement & Implemented control \\
\midrule
Engaging, pleasant imagery can redirect unwanted pre-sleep thought~\cite{harvey2002}. & Capture attention without threat, urgency, or an unresolved goal. & Concrete opening, continuous low-stakes movement, sensory palette, limited dialogue, and bounded novelty. \\
Bedtime-story trials are mixed and the strongest active-controlled result is null~\cite{vazzaz2026}. & Do not represent the system as treatment or guarantee sleep. & Claim boundary in prompts, interface copy, validation rules, paper, and evaluation plan. \\
Daily-listening survey respondents more often reported reusing a book and difficulty falling asleep without audio~\cite{berglund2026}. & Support familiarity without engagement-maximizing habit mechanics. & Recurring-world mode, autoplay off, no streak objective, and reliance-oriented telemetry. \\
Environmental sound can disturb sleep; masking evidence is inconsistent~\cite{capezuti2022,who2009}. & Keep level and dynamics controlled and ambience optional. & Voice-first mix, two-pass loudness normalization, post-encode loudness, loudness-range, and true-peak gate, and timed fade. \\
LLMs violate schemas and numerical length constraints~\cite{lu2025,retkowski2025}. & Verify rather than trust generation instructions. & Parseable front matter, server-computed word and taper budgets, deterministic validators, one targeted retry, then quarantine. \\
TTS realization varies across voices, languages, and providers~\cite{ssml2010,kim2021}. & Separate narrative intent from synthesis configuration and waveform QA. & Provider adapter, paragraph chunks, locale settings, pronunciation metadata, silent join, and post-synthesis measurement. \\
\bottomrule
\end{tabularx}
\end{table*}

\section{System Architecture}
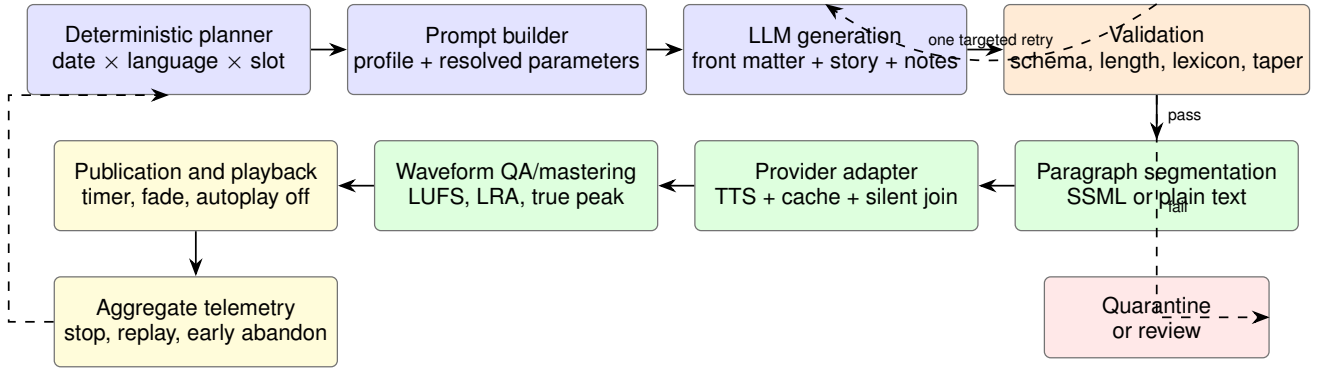
\begin{figure*}[!t]
\centering
\begin{tikzpicture}[
  font=\sffamily\scriptsize,
  node distance=5mm and 4mm,
  stage/.style={draw=black!55, rounded corners=2pt, minimum width=3.15cm,
    minimum height=10mm, align=center, inner sep=2pt},
  plan/.style={stage, fill=blue!11},
  guard/.style={stage, fill=orange!16},
  audio/.style={stage, fill=green!13},
  output/.style={stage, fill=yellow!18},
  side/.style={draw=black!55, rounded corners=2pt, fill=red!9,
    minimum width=2.5cm, minimum height=9mm, align=center, inner sep=2pt},
  arrow/.style={-{Stealth[length=2mm]}, line width=0.55pt},
  feedback/.style={-{Stealth[length=2mm]}, line width=0.55pt, dashed}
]
  \node[plan] (planner) {Deterministic planner\\date $\times$ language $\times$ slot};
  \node[plan, right=of planner] (prompt) {Prompt builder\\profile + resolved parameters};
  \node[plan, right=of prompt] (llm) {LLM generation\\front matter + story + notes};
  \node[guard, right=of llm] (validation) {Validation\\schema, length, lexicon, taper};

  \node[audio, below=of validation] (segment) {Paragraph segmentation\\SSML or plain text};
  \node[audio, left=of segment] (tts) {Provider adapter\\TTS + cache + silent join};
  \node[audio, left=of tts] (master) {Waveform QA/mastering\\LUFS, LRA, true peak};
  \node[output, left=of master] (publish) {Publication and playback\\timer, fade, autoplay off};
  \node[output, below=of publish] (telemetry) {Aggregate telemetry\\stop, replay, early abandon};
  \node[side, below=of segment] (quarantine) {Quarantine\\or review};

  \draw[arrow] (planner) -- (prompt);
  \draw[arrow] (prompt) -- (llm);
  \draw[arrow] (llm) -- (validation);
  \draw[arrow] (validation) -- node[right, font=\sffamily\tiny] {pass} (segment);
  \draw[arrow] (segment) -- (tts);
  \draw[arrow] (tts) -- (master);
  \draw[arrow] (master) -- (publish);
  \draw[arrow] (publish) -- (telemetry);

  \draw[feedback] (validation.north) to[bend left=35]
    node[above, font=\sffamily\tiny] {one targeted retry} (llm.north);
  \draw[feedback] (validation.south) |- node[near start, right, font=\sffamily\tiny] {fail} (quarantine.east);
  \draw[feedback] (telemetry.west) -- ++(-5mm,0) |- (planner.south);
\end{tikzpicture}
\caption{End-to-end production pipeline. Solid arrows show the publication path; dashed arrows show correction, quarantine, and aggregate feedback paths.}
\label{fig:pipeline}
\end{figure*}

The production system executes a nightly sequence from deterministic slot planning to publication and feedback (Fig.~\ref{fig:pipeline}). Generation is deliberately one stage in a larger controlled process.

\subsection{Deterministic planning}
For each run date and language, the planner resolves a slot record before any model call. The record includes target duration, engagement arm, story domain, movement mode, point of view, familiarity mode, setting, sensory palette, weather and time-of-day profile, repetition strategy, and voice configuration. Repeating the same date--language run is idempotent. Variation is therefore scheduled over controlled pools rather than obtained by repeatedly asking a model to be ``creative.'' This prevents accidental local theme collapse and makes each artifact traceable to a resolved input.

The engagement arm is also assigned by the planner. Fixed allocation across low-, medium-, and higher-engagement variants permits later comparative experiments while preventing production traffic from drifting toward whichever content maximizes completion.

\subsection{Prompt contract and duration budget}
The prompt is composed from stable system rules, the sleep-story skill, the resolved slot, an output schema, and optional retry diagnostics. It requests machine-parseable front matter, a title, one continuous narration body, and a short synthesis-notes block. The body must not expose its phase labels, duration budget, or validation state.

For language $\ell$, duration planning can be expressed as
\begin{equation}
W_{\ell}^{\star}=T\frac{r_{s,\ell}}{\bar{s}_{\ell}},
\qquad
W_{\ell}\in[0.93W_{\ell}^{\star},1.07W_{\ell}^{\star}],
\label{eq:duration}
\end{equation}
where $T$ is target minutes, $r_{s,\ell}$ is the configured syllable rate, and $\bar{s}_{\ell}$ is an empirical syllables-per-word estimate. When a language-specific direct word-rate estimate is available, it can replace the ratio. The $\pm7\%$ band is the target range given to the generator; the blocking validator check rejects a script whose syllable-estimated duration deviates from $T$ by more than 8\%. Both are implementation tolerances, not evidence-based sleep thresholds. Languages for which words are a poor duration unit require a different estimator; Japanese mora-level validation remains an identified implementation gap.

\subsection{Four-phase attention taper}
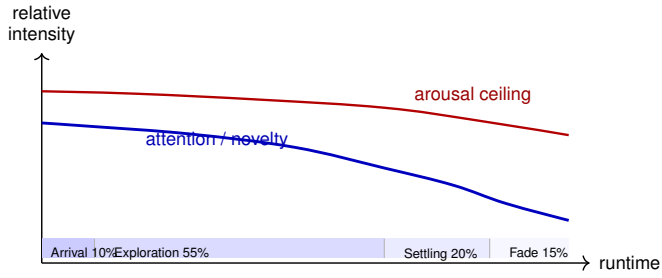
\begin{figure}[!t]
\centering
\begin{tikzpicture}[x=0.071cm,y=0.82cm,font=\sffamily\scriptsize]
  \draw[->, line width=0.55pt] (0,0) -- (104,0) node[right] {runtime};
  \draw[->, line width=0.55pt] (0,0) -- (0,3.45) node[above, align=center] {relative\\intensity};

  \fill[blue!20] (0,0.08) rectangle (10,0.40);
  \fill[blue!14] (10,0.08) rectangle (65,0.40);
  \fill[blue!8]  (65,0.08) rectangle (85,0.40);
  \fill[blue!3]  (85,0.08) rectangle (100,0.40);
  \draw[black!30] (10,0.08) -- (10,0.40);
  \draw[black!30] (65,0.08) -- (65,0.40);
  \draw[black!30] (85,0.08) -- (85,0.40);
  \node[anchor=north west, font=\sffamily\tiny] at (0,0.40) {Arrival 10\%};
  \node[anchor=north west, font=\sffamily\tiny] at (12,0.40) {Exploration 55\%};
  \node[anchor=north west, font=\sffamily\tiny] at (67,0.40) {Settling 20\%};
  \node[anchor=north west, font=\sffamily\tiny] at (87,0.40) {Fade 15\%};

  \draw[red!70!black, line width=0.9pt]
    plot[smooth, tension=0.7] coordinates
    {(0,2.82) (18,2.78) (42,2.68) (65,2.55) (82,2.35) (100,2.10)};
  \node[red!60!black, anchor=west] at (69,2.76) {arousal ceiling};

  \draw[blue!75!black, line width=1.1pt]
    plot[smooth, tension=0.65] coordinates
    {(0,2.30) (10,2.24) (28,2.12) (48,1.90) (65,1.56) (78,1.28) (88,0.98) (100,0.70)};
  \node[blue!70!black, anchor=west] at (18,2.02) {attention / novelty};
\end{tikzpicture}
\caption{Conceptual attention taper. Attentional demand is sufficient to redirect thought early, then declines; the permitted arousal band remains low throughout. The curves are design targets, not measured physiology.}
\label{fig:attention}
\end{figure}

The narrative contract partitions the script into Arrival (10\%), Exploration (55\%), Settling (20\%), and Fade (15\%). The unit is the word index of the script, not measured playback time: for a target of $W$ words, the contract places the boundaries at $\mathrm{round}(0.10W)$, $\mathrm{round}(0.65W)$, and $\mathrm{round}(0.85W)$. After generation, the server recomputes the Settling--Fade boundary from the realized word count $W_r$ as $w_f=\mathrm{round}\big(W_r(1-T_f/T)\big)$, where the fade duration $T_f$ is 4~min for stories of 15~min or longer and $\max(2,\lfloor T/4\rfloor)$~min otherwise; $T_f/T$ equals the nominal 15\% at about 25~min. The configured synthesis rate is constant within a story, but sentence and paragraph pauses lengthen after $w_f$ (for English, from 800 to 1000~ms and from 1600 to 2200~ms; other locales use slightly longer values), and providers that accept a style instruction are asked to slow slightly toward the end. Word-index boundaries therefore do not map linearly to playback time, and Fade occupies a larger share of playback time than of words. Phase boundaries are not yet located in the rendered waveform; forced alignment of text to audio would allow this and is future work. Arrival immediately provides a concrete safe place, object, or activity. Exploration introduces benign descriptive discoveries along one simple thread. Settling decreases the rate of new entities and scene changes. Fade introduces no important names, goals, locations, or questions and ends without a cliffhanger. Figure~\ref{fig:attention} shows the conceptual profile.

The system seeks curiosity about \emph{what will be described next}, not concern about \emph{what will happen}. Hard prompt constraints exclude consequential stakes, open mysteries, direct sleep instructions, body-state claims, exclamations, and direct questions. Movement modes---journey, process, cycle, catalogue, or tending---produce variety while retaining a low-consequence arc.

\subsection{Guarded generation and bounded retry}
Validation proceeds in layers, and each check is either blocking (an error) or advisory (a warning recorded with the story). The parser first reads the front matter and narration body. Blocking checks reject a syllable-estimated duration more than 8\% from target, any question mark or exclamation mark, any hit in per-language lists of banned high-arousal and sleep-instruction terms, anticipation phrasing, listener self-address, or meta-reference to the story, and a closing gesture that resolves the narrative. Advisory checks report word-count deviation above 7\%, median sentence length outside 12--22 words, an abstraction-rate proxy, and two taper checks after the Fade boundary $w_f$: the new-word rate relative to the second half of the body, and mean sentence length relative to the body. Taper checks remain advisory because rule-based proxies cannot reliably infer narrative stakes; these warnings are reviewed by an operator rather than used to reject automatically.

If validation fails, one retry is issued with the original resolved slot and targeted diagnostics. The retry asks for correction of named violations rather than unconstrained regeneration. A second failure is quarantined for review. This limit bounds cost and prevents a failing slot from silently entering an open-ended generation loop.

\subsection{Synthesis, mastering, and delivery}
Accepted scripts are divided at paragraph boundaries and passed through a provider adapter supporting SSML or plain text. The adapter resolves locale, voice, pronunciation substitutions, rate, and inter-paragraph pauses. Chunks are cached and joined with generated silence so a transient failure does not require resynthesizing the entire story.

The joined waveform is resampled to 44.1~kHz mono and mastered with FFmpeg's \texttt{loudnorm} filter, which measures according to EBU R~128~\cite{ebu128}. The sequence is: (i) a measurement pass on the lossless source; (ii) a normalization pass that uses the measured values, with targets of $-22$~LUFS integrated loudness, $-3$~dBTP true peak, and 4~LU loudness range, followed by a 150~s fade-out; (iii) AAC encoding at 96~kb/s; and (iv) a second measurement of the saved, encoded file. Publication requires integrated loudness in $[-26,-18]$~LUFS, true peak $\le-2.5$~dBTP, and loudness range $\le4.5$~LU on that saved file. If it fails, mastering restarts from the lossless source with dynamic-range compression of ratio 2:1 and then 4:1, for at most three attempts. The limits are never relaxed, and a file that still fails is withheld from publication. The application disables automatic continuation, offers a sleep timer, and supports screen-off playback. These limits fix level relative to digital full scale; device volume remains under user control, so they cannot guarantee the sound-pressure level at the pillow.

\section{Implementation and Guardrail Layers}
The prototype separates configuration from generated artifacts. A slot record identifies the planner version, language, content parameters, prompt version, model/provider identifiers, validation results, retry state, synthesis settings, and audio measurements. This separation permits a script to be regenerated or resynthesized without losing provenance. Recording why each artifact was accepted, flagged, or quarantined follows a human-centered view of explainable AI, in which automated decisions remain inspectable by the people who oversee them~\cite{makridis2026xai}.

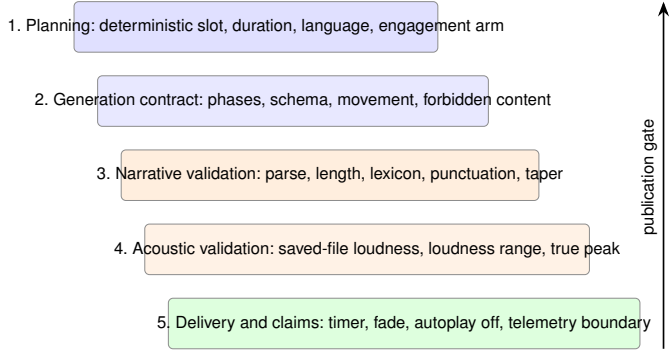
\begin{figure}[!t]
\centering
\begin{tikzpicture}[font=\sffamily\scriptsize]
  \filldraw[draw=black!45, fill=blue!12, rounded corners=2pt]
    (0,4.8) rectangle (5.4,5.55);
  \node[align=center] at (2.7,5.175) {1. Planning: deterministic slot, duration, language, engagement arm};

  \filldraw[draw=black!45, fill=blue!9, rounded corners=2pt]
    (0.35,3.7) rectangle (6.15,4.45);
  \node[align=center] at (3.25,4.075) {2. Generation contract: phases, schema, movement, forbidden content};

  \filldraw[draw=black!45, fill=orange!13, rounded corners=2pt]
    (0.7,2.6) rectangle (6.9,3.35);
  \node[align=center] at (3.8,2.975) {3. Narrative validation: parse, length, lexicon, punctuation, taper};

  \filldraw[draw=black!45, fill=orange!10, rounded corners=2pt]
    (1.05,1.5) rectangle (7.65,2.25);
  \node[align=center] at (4.35,1.875) {4. Acoustic validation: saved-file loudness, loudness range, true peak};

  \filldraw[draw=black!45, fill=green!13, rounded corners=2pt]
    (1.4,0.4) rectangle (8.4,1.15);
  \node[align=center] at (4.9,0.775) {5. Delivery and claims: timer, fade, autoplay off, telemetry boundary};

  \draw[-{Stealth[length=2.2mm]}, line width=0.7pt] (8.75,0.4) -- (8.75,5.55);
  \node[rotate=90, anchor=south] at (8.78,2.95) {publication gate};
\end{tikzpicture}
\caption{Defense-in-depth guardrails. Publication requires conformance at every layer.}
\label{fig:guardrails}
\end{figure}

Figure~\ref{fig:guardrails} groups controls into five layers. A failure at a lower layer cannot be compensated by fluent text at a higher layer: a well-written script with an invalid duration, an unsafe claim, or a clipped waveform remains unpublishable.

\subsection{Multilingual operation}
Language is a first-class planning key rather than a final translation step. Each locale can define pools, sentence-complexity limits, duration coefficients, banned expressions, pronunciation rules, voices, and punctuation constraints. This design avoids assuming that an English word budget or prosody profile transfers directly to other languages. Current limitations include incomplete mora-based duration validation for Japanese and the absence of a completed cross-language perceptual study. Multilingual support should therefore be understood as an implemented architecture with language-specific calibration still in progress.

\subsection{Current implementation boundary}
We separate four levels of status. \emph{Implemented and unit-tested:} the duration and taper metadata, the multilingual validator, truncation handling for model output, SSML construction, lexical paragraph repair, and the saved-file mastering gate. \emph{Implemented:} planning, the prompt contract, retry and quarantine, provider adapters, and publication metadata. \emph{Deployed:} the pipeline has published stories in English, German, French, Hindi, and Japanese to the production catalogue. \emph{Designed but not implemented:} mora-based Japanese duration validation, phase timing measured in rendered audio, a cross-language perceptual study, and the controlled study of Section~\ref{sec:evaluation}. Apart from the five-story pilot in Section~\ref{sec:evaluation}, this paper does not report conformance results on a fixed corpus, production-scale reliability statistics, or clinical outcome data; the other evaluation measures below are a plan, not results. Those omissions define the present evaluation boundary rather than being treated as positive results.

\section{Evaluation Framework}
\label{sec:evaluation}
Evaluation is divided into technical conformance, behavioral telemetry, and controlled sleep outcomes. Mixing these levels would invite an efficacy claim from an engineering metric.

\subsection{Technical conformance}
The first evaluation level can be run without users. Recommended measures include first-pass schema validity, first-pass acceptance, retry recovery, quarantine rate, absolute word-count error, phase-budget error, banned-cue precision and recall on an annotated test set, late-story entity-introduction rate, synthesis failure rate, duration error, integrated loudness, loudness range, maximum true peak, and boundary-discontinuity counts. Results should be reported by language, voice, generation model, and story domain. A fixed regression corpus is necessary to detect prompt or provider changes.

\subsection{Pilot measurement}
As an initial check, Table~\ref{tab:pilot} reports the five stories published from one nightly run (12 October 2026): one 20-min slot each in English, German, French, Hindi, and Japanese, all voiced with the same ElevenLabs voice. All five passed the blocking script checks with no warnings and passed the saved-file gate. Integrated loudness was within 0.3~LU of the $-22$~LUFS target. True peak and loudness range met the acceptance limits but sat slightly above the mastering targets of $-3$~dBTP (all five) and 4~LU (four of five). Rendered duration, which no gate checks, was within 3\% of target for English and German but 13\%, 24\%, and 38\% short for French, Japanese, and Hindi. The syllables-per-word estimate in~\eqref{eq:duration} therefore overestimates speaking time for these languages with this voice; a script can pass the 8\% check and still render far from its target. Five stories from a single run are a pilot, not a fixed-corpus evaluation, and published records cannot show first-pass acceptance, retry, or quarantine rates.

\begin{table}[!t]
\caption{Pilot Measurement: Five Published Stories (20-min Target)}
\label{tab:pilot}
\centering
\scriptsize
\begin{tabular}{lrrrrr}
\toprule
Lang. & Rendered (min) & Error & LUFS & TP (dBTP) & LRA (LU) \\
\midrule
en & 20.4 & $+2.0\%$ & $-22.22$ & $-2.68$ & 4.4 \\
de & 20.6 & $+2.9\%$ & $-22.23$ & $-2.81$ & 4.2 \\
fr & 17.4 & $-13.0\%$ & $-22.28$ & $-2.90$ & 4.4 \\
hi & 12.4 & $-38.2\%$ & $-22.17$ & $-2.98$ & 3.0 \\
ja & 15.2 & $-24.1\%$ & $-22.20$ & $-2.98$ & 4.4 \\
\bottomrule
\end{tabular}
\end{table}

\subsection{Behavioral telemetry}
Let $T_a$ denote audio duration and $t_s$ the final playhead position in a nocturnal session. The system records a normalized stop fraction
\begin{equation}
f_s=\min\left(\frac{t_s}{T_a},1\right),
\label{eq:stop-fraction}
\end{equation}
along with a reached-end indicator, early abandonment before 120~s, and same-night replay. These metrics support content-fit diagnosis, but none identifies sleep. A low $f_s$ may indicate sleep, rejection, interruption, or technical failure; a high $f_s$ may indicate wakefulness, enjoyment, or a short timer. Accordingly, aggregate telemetry is used to generate hypotheses and select material for trials, not to label an individual as asleep.

The system also avoids a completion-maximizing objective and monitors repeated nightly use as a possible reliance signal. No streak reward or escalating notification should be attached to bedtime listening. Telemetry collection should be minimized, consented, access-controlled, and retained only as long as required for the stated evaluation.

\subsection{Prospective controlled study}
A preregistered at-home study should compare: (A) generated stories with the four-phase taper, (B) duration- and narrator-matched stories without the taper constraints, and (C) a digital control with the same interface but no narrated content. A within-person randomized design can reduce between-person variability. The primary endpoint should be sleep-onset latency measured with validated home EEG where feasible; actigraphy alone can misclassify quiet wakefulness. Secondary endpoints should include wake after sleep onset, total sleep time, sleep efficiency, pre-sleep cognitive arousal, morning restoration, next-day sleepiness, adverse events, and perceived reliance.

A factorial extension can test engagement level, 15 versus 30 minutes, voice-only versus optional ambience, and unfamiliar versus recurring-world content. The success criterion is not higher completion. It is shorter sleep-onset latency without increased awakenings, next-day impairment, or reliance. Analyses should report average effects and responder distributions because personalized audio may benefit some listeners while disturbing others.

\section{Safety, Ethics, and Limitations}
The system is a relaxation product, not CBT-I or a diagnostic device. Interface and narration text must not promise treatment, instruct users to ignore persistent symptoms, or imply that a specific frequency changes brain rhythms. Users reporting chronic insomnia, marked daytime impairment, loud snoring, witnessed breathing pauses, or gasping should be directed toward qualified care.

Dependence is a separate concern from acute safety. The observational association between frequent listening and difficulty sleeping without audio does not establish causation~\cite{berglund2026}, but it supports conservative product mechanics and explicit measurement. Privacy is also material: nighttime playback patterns can reveal routines and health-related behavior. The architecture should avoid microphone capture by default and separate operational logs from research datasets.

Several limitations remain. First, the attention taper is an evidence-informed design hypothesis, not a validated dose. Second, the literature transfers findings from imagery, music, noise, and clinical settings to generated narration. Third, lexical and structural rules cannot guarantee low emotional arousal, especially across cultures and languages. Fourth, TTS voices and consumer playback devices introduce uncontrolled prosodic and acoustic variation. Fifth, stop-position telemetry is intrinsically ambiguous. Finally, the present work reports architecture and implementation status rather than clinical effectiveness or comparative generation-quality results.

\section{Conclusion}
We presented an evidence-grounded pipeline for AI-generated, narrated sleep stories. Its central design decision is to optimize neither flatness nor ordinary engagement, but sufficient attentional capture under a low-arousal constraint followed by a controlled taper. Deterministic planning, explicit prompt contracts, external validation with stated thresholds, bounded retry, provider-independent synthesis, saved-file loudness checks, and non-completion-centric telemetry make each published artifact traceable to its inputs and checks. The architecture creates a credible platform for controlled evaluation; it does not itself demonstrate improved sleep. Future work should publish language-stratified technical results and a preregistered trial using objective and subjective sleep outcomes.

\bibliographystyle{IEEEtran}
\bibliography{references}

@article{harmat2008,
  author  = {Harmat, L{\'a}szl{\'o} and Tak{\'a}cs, Johanna and B{\'o}dizs, R{\'o}bert},
  title   = {Music Improves Sleep Quality in Students},
  journal = {Journal of Advanced Nursing},
  volume  = {62},
  number  = {3},
  pages   = {327--335},
  year    = {2008},
  doi     = {10.1111/j.1365-2648.2008.04602.x}
}

@article{kirk2022,
  author  = {Kirk, Ulrich and Ngnoumen, Christelle and Clausel, Alicia and Purvis, Clare Kennedy},
  title   = {Using Actigraphy and Heart Rate Variability ({HRV}) to Assess Sleep Quality and Sleep Arousal of Three App-Based Interventions: Sleep Music, Sleepcasts, and Guided Mindfulness},
  journal = {Journal of Cognitive Enhancement},
  volume  = {6},
  number  = {2},
  pages   = {216--231},
  year    = {2022},
  doi     = {10.1007/s41465-021-00233-4}
}

@article{economides2023,
  author  = {Economides, Marcos and Male, Rhian and Bolton, Heather and Cavanagh, Kate},
  title   = {Feasibility and Preliminary Efficacy of App-Based Audio Tools to Improve Sleep Health in Working Adults Experiencing Poor Sleep: A Multi-Arm Randomized Pilot Trial},
  journal = {Sleep},
  volume  = {46},
  number  = {7},
  pages   = {zsad053},
  year    = {2023},
  doi     = {10.1093/sleep/zsad053}
}

@article{mermer2024,
  author  = {Mermer, Emine and Arslan, Sevilay},
  title   = {The Effect of Audiobooks on Sleep Quality and Vital Signs in Intensive Care Patients},
  journal = {Intensive and Critical Care Nursing},
  volume  = {80},
  pages   = {103552},
  year    = {2024},
  doi     = {10.1016/j.iccn.2023.103552}
}

@article{vazzaz2026,
  author  = {Vazzaz, Jessica and Matcham, Faith and Economides, Marcos and Bolton, Heather and Cavanagh, Kate},
  title   = {{USleep}: Efficacy of App-Based Audio Interventions to Improve Sleep Disturbance in Working Adults, a Multi-Arm Randomized Controlled Trial},
  journal = {Sleep},
  volume  = {49},
  number  = {9},
  pages   = {zsag154},
  year    = {2026},
  doi     = {10.1093/sleep/zsag154}
}

@article{berglund2026,
  author  = {Berglund, Karl and Meth, Elisa M. S. and Xue, Peiying and Benedict, Christian},
  title   = {Bedtime Audiobook Listening Motives and Insomnia Status: Survey Study},
  journal = {JMIR Formative Research},
  volume  = {10},
  pages   = {e95064},
  year    = {2026},
  doi     = {10.2196/95064}
}

@misc{nedimo,
  author       = {{Nedimo Team}},
  title        = {Nedimo Web Application and Project Website},
  year         = {2026},
  howpublished = {Online},
  url          = {https://nedimo.com},
  note         = {Accessed: 2026-10-29}
}

@misc{nedimo-play,
  author       = {{Nedimo Team}},
  title        = {Nedimo: Sleep Stories \& Sounds, Android App on Google Play},
  year         = {2026},
  howpublished = {Online},
  url          = {https://play.google.com/store/apps/details?id=com.nedimo},
  note         = {Accessed: 2026-10-29}
}

@misc{nedimo-appstore,
  author       = {{Nedimo Team}},
  title        = {Nedimo: Sleep Stories, iOS App on the App Store},
  year         = {2026},
  howpublished = {Online},
  url          = {https://apps.apple.com/app/id6803620912},
  note         = {Accessed: 2026-10-29}
}

@article{harvey2002,
  author  = {Harvey, Allison G. and Payne, Suzanna},
  title   = {The Management of Unwanted Pre-Sleep Thoughts in Insomnia: Distraction with Imagery versus General Distraction},
  journal = {Behaviour Research and Therapy},
  volume  = {40},
  number  = {3},
  pages   = {267--277},
  year    = {2002},
  doi     = {10.1016/S0005-7967(01)00012-2}
}

@article{jespersen2022,
  author  = {Jespersen, Kira V. and Pando-Naude, Victor and Koenig, Julian and Jennum, Poul and Vuust, Peter},
  title   = {Listening to Music for Insomnia in Adults},
  journal = {Cochrane Database of Systematic Reviews},
  number  = {8},
  pages   = {CD010459},
  year    = {2022},
  doi     = {10.1002/14651858.CD010459.pub3}
}

@article{scarratt2023,
  author  = {Scarratt, Rebecca Jane and Heggli, Ole Adrian and Vuust, Peter and Jespersen, Kira Vibe},
  title   = {The Audio Features of Sleep Music: Universal and Subgroup Characteristics},
  journal = {PLOS ONE},
  volume  = {18},
  number  = {1},
  pages   = {e0278813},
  year    = {2023},
  doi     = {10.1371/journal.pone.0278813}
}

@article{capezuti2022,
  author  = {Capezuti, Elizabeth and Pain, Kevin and Alamag, Evelyn and Chen, Xinqing and Philibert, Valicia and Krieger, Ana C.},
  title   = {Systematic Review: Auditory Stimulation and Sleep},
  journal = {Journal of Clinical Sleep Medicine},
  volume  = {18},
  number  = {6},
  pages   = {1697--1709},
  year    = {2022},
  doi     = {10.5664/jcsm.9860}
}

@techreport{who2009,
  author      = {{World Health Organization Regional Office for Europe}},
  title       = {Night Noise Guidelines for Europe},
  institution = {World Health Organization},
  address     = {Copenhagen, Denmark},
  year        = {2009},
  isbn        = {978-92-890-4173-7},
  url         = {https://www.who.int/europe/publications/i/item/9789289041737}
}

@article{saskovets2025,
  author  = {Saskovets, Marina and Lohachov, Mykhailo and Liang, Zilu},
  title   = {Paralanguage as a Tool for Shaping Stress Response in Listeners: Multimodal Physiological Sensing Study},
  journal = {Comprehensive Psychoneuroendocrinology},
  volume  = {23},
  pages   = {100309},
  year    = {2025},
  doi     = {10.1016/j.cpnec.2025.100309}
}

@article{chang2015,
  author  = {Chang, Anne-Marie and Aeschbach, Daniel and Duffy, Jeanne F. and Czeisler, Charles A.},
  title   = {Evening Use of Light-Emitting eReaders Negatively Affects Sleep, Circadian Timing, and Next-Morning Alertness},
  journal = {Proceedings of the National Academy of Sciences},
  volume  = {112},
  number  = {4},
  pages   = {1232--1237},
  year    = {2015},
  doi     = {10.1073/pnas.1418490112}
}

@inproceedings{lu2025,
  author    = {Lu, Yaxi and Li, Haolun and Cong, Xin and Zhang, Zhong and Wu, Yesai and Lin, Yankai and Liu, Zhiyuan and Liu, Fangming and Sun, Maosong},
  title     = {Learning to Generate Structured Output with Schema Reinforcement Learning},
  booktitle = {Proceedings of the 63rd Annual Meeting of the Association for Computational Linguistics},
  pages     = {4905--4918},
  publisher = {Association for Computational Linguistics},
  year      = {2025},
  doi       = {10.18653/v1/2025.acl-long.243}
}

@misc{makridis2024fairyland,
  author       = {Makridis, Georgios and Oikonomou, Athanasios and Koukos, Vasileios},
  title        = {{FairyLandAI}: Personalized Fairy Tales Utilizing {ChatGPT} and {DALLE-3}},
  howpublished = {arXiv preprint arXiv:2407.09467},
  year         = {2024},
  url          = {https://arxiv.org/abs/2407.09467}
}

@incollection{makridis2026xai,
  author    = {Makridis, Georgios},
  title     = {Human-Centered Explainable {AI} in Education},
  booktitle = {Explainable Artificial Intelligence and Interpretable Machine Learning in Education},
  editor    = {Khine, Myint Swe},
  publisher = {CRC Press},
  pages     = {58--73},
  year      = {2026},
  doi       = {10.1201/9781003713371-4}
}

@inproceedings{retkowski2025,
  author    = {Retkowski, Fabian and Waibel, Alexander},
  title     = {Zero-Shot Strategies for Length-Controllable Summarization},
  booktitle = {Findings of the Association for Computational Linguistics: NAACL 2025},
  pages     = {551--572},
  address   = {Albuquerque, New Mexico},
  publisher = {Association for Computational Linguistics},
  year      = {2025},
  doi       = {10.18653/v1/2025.findings-naacl.34}
}

@techreport{ssml2010,
  author      = {{W3C Voice Browser Working Group}},
  title       = {Speech Synthesis Markup Language ({SSML}) Version 1.1},
  institution = {World Wide Web Consortium},
  type        = {W3C Recommendation},
  year        = {2010},
  url         = {https://www.w3.org/TR/speech-synthesis11/}
}

@inproceedings{kim2021,
  author    = {Kim, Jaehyeon and Kong, Jungil and Son, Juhee},
  title     = {Conditional Variational Autoencoder with Adversarial Learning for End-to-End Text-to-Speech},
  booktitle = {Proceedings of the 38th International Conference on Machine Learning},
  series    = {Proceedings of Machine Learning Research},
  volume    = {139},
  pages     = {5530--5540},
  year      = {2021},
  url       = {https://proceedings.mlr.press/v139/kim21f.html}
}

@techreport{ebu128,
  author      = {{European Broadcasting Union}},
  title       = {Loudness Normalisation and Permitted Maximum Level of Audio Signals},
  institution = {European Broadcasting Union},
  number      = {EBU R 128, Version 5.0},
  year        = {2023},
  url         = {https://tech.ebu.ch/publications/r128}
}

\end{document}